\documentclass[conference]{IEEEtran}
\IEEEoverridecommandlockouts

\usepackage{graphicx}
\usepackage{amsmath,amssymb}
\usepackage{cite}
\usepackage{xcolor}
\usepackage{url}
\usepackage{soul}
\usepackage[colorlinks=true, linkcolor=blue, citecolor=violet, urlcolor=cyan]{hyperref}

\title{Utilizing Vision-Language Models as Action Models for Intent Recognition and Assistance \\ (Extended Abstract)
\vspace{-0.7em}}

\author{\IEEEauthorblockN{Cesar Alan Contreras$^1$, Manolis Chiou$^2$, Alireza Rastegarpanah$^3$, Michal Szulik$^4$, Rustam Stolkin$^1$}
\IEEEauthorblockA{$^1$ School of Metallurgy \& Materials, University of Birmingham, Birmingham B15 2SE, United Kingdom\\
$^2$ School of Electronic Engineering and Computer Science, Queen Mary University of London, London E14 4NS, United Kingdom\\
$^3$ School of Computer Science and Digital Technologies, Aston University, Birmingham B4 7ET, United Kingdom\\
$^4$ United Kingdom National Nuclear Laboratory Ltd., Warrington WA3 6AE, United Kingdom}
\thanks{Corresponding author: Cesar Alan Contreras (email: cac214@bham.ac.uk).}
\thanks{This work is funded by the Nuclear Decommissioning Authority (NDA) and supported by the United Kingdom National Nuclear Laboratory (UKNNL). In addition, it was supported by 
the UK Research and Innovation (UKRI) project “REBELION” under Grant 101104241.}
\vspace{-2.5em}
}

\begin{document}

\maketitle

% ---------------------------------------
% Abstract
% ---------------------------------------
\begin{abstract}
Human-robot collaboration requires robots to quickly infer user intent, provide transparent reasoning, and assist users in achieving their goals. Our recent work introduced GUIDER, our framework for inferring navigation and manipulation intents.  We propose augmenting GUIDER with a vision-language model (VLM) and a text-only language model (LLM) to form a semantic prior that filters objects and locations based on the mission prompt.
A vision pipeline (YOLO for object detection and the Segment Anything Model for instance segmentation) feeds candidate object crops into the VLM, which scores their relevance given an operator prompt; in addition, the list of detected object labels is ranked by a text-only LLM. These scores weight the existing navigation and manipulation layers of GUIDER, selecting context-relevant targets while suppressing unrelated objects. Once the combined belief exceeds a threshold, autonomy changes occur, enabling the robot to navigate to the desired area and retrieve the desired object, while adapting to any changes in the operator's intent.  Future work will evaluate the system on Isaac Sim using a Franka Emika arm on a Ridgeback base, with a focus on real-time assistance.
\end{abstract}

\begin{IEEEkeywords}
Intent inference, assistance, vision-language models, human-robot interaction, mobile manipulation, probabilistic reasoning
\vspace{-1.4em}
\end{IEEEkeywords}

% ---------------------------------------
\section{Introduction}
Robots intended for collaborative work must understand human goals to assist effectively, failing to recognise intent can lead to conflicting behaviours and increase cognitive load on users.  Previously, we introduced \emph{GUIDER} \cite{contreras2025probabilistic}, a probabilistic framework coupling a navigation and manipulation layer.  The navigation layer used controller inputs, an occupancy map, and a synergy map to maintain beliefs over areas of interest, while the manipulation layer fused visual saliency, instance segmentation, and grasp feasibility to estimate object intent. In the trials, GUIDER achieved similar prediction times compared to the baselines, and improved on stability, achieving a stability of 93-100\%. Although GUIDER provided accurate real-time estimates of operator intent, it is a predictive tool that still requires the human operator to manually command the robot to complete the task.  Closing the loop between inference and assistance is necessary for reducing cognitive load and enabling fluent collaboration.

Recent work links language and perception to robot behaviour along three themes. First, instruction grounding: language is tied to skills or value functions so robots follow high-level commands with task priors \cite{ahn2022saycan}. Second, vision-language action: VLMs and related models map images and text to action abstractions, enabling generalisation to novel objects and affordances \cite{pmlr-v229-zitkovich23a, gao2024physobjects}. Third, proactive intent modelling: language is used to track human activity, reason over tasks, or inject structured knowledge for interpretable predictions and on-the-fly procedures \cite{huang2024lit, zhou2024lkirf, liu2024vlm}. 
These studies show strong open-vocabulary reasoning, but two gaps remain for human-in-the-loop mobile manipulation: (i) fusing semantic context as an explicit probability factor with continuous intent beliefs over both navigation areas and candidate objects, and (ii) closing the loop from intent inference to assistance with prompt-conditioned commitment policies. Our work addresses both.

We aim to utilize LLMs to aid or assist human operators, as shown in the literature, while improving their dynamic integration and assistance. For this purpose, in this study, we propose exploring whether a VLM with an optional text-only LLM can be integrated into our dual-phase framework as an additional probability metric, also using it to support operator actions. The semantic model filters context-relevant objects and location categories, which can provide a weighting that is fused with the existing navigation and manipulation belief updates in GUIDER.  By conditioning on the mission context, we reduce the search space, enrich the probabilistic model, and bridge the gap between intent prediction and assistance within an integrated reasoning and action model, while explicitly modelling the operator's context through natural language prompts or mission briefs that can be updated adaptively. We refer to the VLM/LLM outputs collectively as a semantic prior that is fused with GUIDER’s beliefs.

\vspace{-0.5em}
% ---------------------------------------
\section{Method / System}
\begin{figure}[tb]
  \centering
  \includegraphics[width=0.99\linewidth]{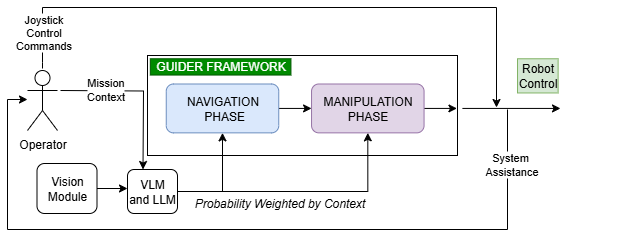}
  \caption{Overview of the VLM/LLM-weighted GUIDER pipeline. A vision module detects and segments objects; each instance provides a class label and an image crop. The mission prompt initialises context.}
  \label{fig:pipeline}
  \vspace{-1.6em}
\end{figure}

Figure \ref{fig:pipeline} illustrates the proposed intent recognition system. An image from the robot's onboard camera is processed by a vision module consisting of YOLO \cite{ultralytics2023yolov8} for object detection and the Segment Anything Model (SAM) \cite{kirillov2023segment} for instance segmentation, where each detected object yields a class label.  At the beginning of the mission, the operator provides a mission prompt defining the context (e.g, “\textbf{Please hand me the television remote}"). Then, a pre-trained VLM takes an image crop from the vision module and a mission prompt from the user, and returns a likelihood that the object matches the described goal. To further condition the model on the mission context, we create a textual prompt that lists the objects and asks the model to rank their relevance.  The resulting scores are normalised and then fused with the evolving likelihoods from GUIDER (both navigation and manipulation).  In effect, the semantic model provides an additional prior that suppresses objects and areas unrelated to the mission. Objects with very low combined scores are pruned to reduce computational load.  The operator can modify the mission prompt at any time to reflect evolving goals, and the resulting weights are immediately reflected in GUIDER's belief update.

\begin{figure}[tb]
  \centering
  \includegraphics[width=0.85\linewidth]{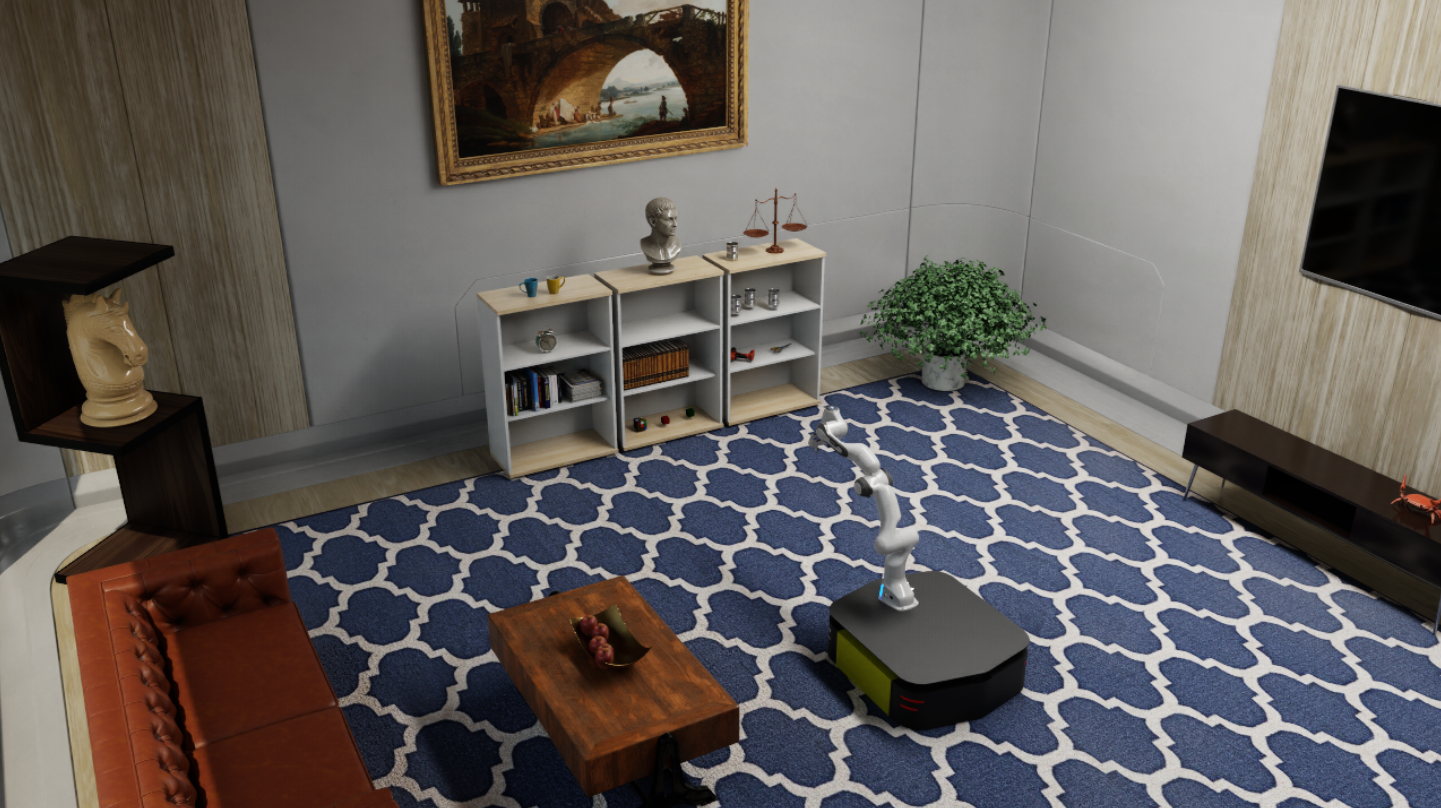}
  \caption{Simulated living-room environment used for planned experiments. The mobile manipulator must interpret mission prompts (e.g., “\textbf{help me in getting some food}") and select relevant objects while ignoring irrelevant items.}
  \label{fig:env}
  \vspace{-1.6em}
\end{figure}

To test the approach, a domestic living room environment is simulated in Isaac Sim (Fig. \ref{fig:env}), where a Franka Emika Panda arm mounted on a Clearpath Ridgeback mobile base receives mission prompts to aid in various tasks. The mission prompt can range from \textbf{specific} (“\textit{Bring me the red mug}”) to \textbf{categorical} (“\textit{Pick up a drink}”), to \textbf{relational} (“\textit{Fetch the cup next to the laptop.}”, “\textit{Anything on the kitchen counter}”).  Detected objects irrelevant to the mission are suppressed by the semantic prior, allowing the system to focus on candidate targets.  Navigation and manipulation phases run in a closed loop; sensor and state feedback continuously update beliefs during assistance.  When the combined probability from GUIDER's navigation and manipulation layers, weighted by the semantic model, exceeds a confidence threshold, the robot transitions from inference to assistance by invoking a shared-autonomy controller or an autonomous mode. In autonomous mode, the mobile base navigates toward the selected object, and the arm executes a grasp. In shared-autonomy mode, controllers reconfigure control axes to centre on the goal.

\vspace{-0.8em}
% ---------------------------------------
\section{Planned Study}
We plan to evaluate the VLM-based GUIDER on a suite of tasks in a living room environment in Isaac Sim.  Each trial will begin with the mobile manipulator at a randomised starting pose in the living room, with the user providing a specific mission prompt or a target at the start of each trial. The robot will autonomously perform a quick scan of the environment and update its belief state, allowing the operator to take control at any time, which causes updates in the intent inference. The robot will be able to commit to autonomy or assistance when the operator accepts the system's suggestion and the probability threshold has been exceeded. 
%When the probability of a particular object exceeds a threshold and is accepted by the user, the robot will commit to autonomy: the base navigates to the object, the arm grasps it, and delivers it to the user without further teleoperation.  
The evaluation metrics will include: (1) time to confident prediction, measured from start until the belief exceeds the threshold; (2) intent accuracy, the fraction of trials where the highest probability object matches the target; and (3) assistance completion time, the duration from prediction to reaching the object or area.  We will compare this new method with the original GUIDER \cite{contreras2025probabilistic} as the baseline, with assistance added.  Experiments will span several object categories (food items, toys, decorations, and tools) and specific, categorical, and relational prompts.
\vspace{-0.6em}
% ---------------------------------------

\section{Contribution and Next Steps}
Our contributions are: (i) Formulate a semantic prior from a VLM (image+text) and an optional LLM (text-only) and fuse it as an explicit probability factor within GUIDER’s navigation and manipulation beliefs; (ii) definition of a commitment rule that triggers shared-autonomy or autonomous assistance when a threshold is crossed; (iii) outline a real-time evaluation on a mobile manipulator with interactive prompt updates. We do not fine-tune models; we exploit zero-shot, open-vocabulary semantics to reduce search and improve transparency.

This extended abstract proposes an intent recognition and action assistance framework that integrates vision-language as an additional probability metric feeding into GUIDER.  By conditioning on the mission prompt and directly computing the relevance of detected objects and locations, the semantic model provides context filtering and generates weights that are fused with GUIDER's navigation and manipulation belief layers. Once the fused belief exceeds a confidence threshold, an assistive controller is invoked to reach the intended area or object. 
The system is designed for domestic mobile manipulation, but the architecture is general and can be applied to other domains.  Future work will perform a comprehensive evaluation against our prior dual-phase framework and baseline approaches in simulation.  We also plan to deploy the method on physical hardware and assess robustness to real-world lighting and occlusion. Finally, we aim to study user interaction by allowing operators to modify mission prompts on the fly, evaluate the system's responsiveness, and investigate how long-horizon sequences and continuous prompts affect both inference and assistance.

\bibliographystyle{IEEEtran}
\bibliography{references}

\begin{thebibliography}{1}
\providecommand{\url}[1]{#1}
\csname url@samestyle\endcsname
\providecommand{\newblock}{\relax}
\providecommand{\bibinfo}[2]{#2}
\providecommand{\BIBentrySTDinterwordspacing}{\spaceskip=0pt\relax}
\providecommand{\BIBentryALTinterwordstretchfactor}{4}
\providecommand{\BIBentryALTinterwordspacing}{\spaceskip=\fontdimen2\font plus
\BIBentryALTinterwordstretchfactor\fontdimen3\font minus \fontdimen4\font\relax}
\providecommand{\BIBforeignlanguage}[2]{{%
\expandafter\ifx\csname l@#1\endcsname\relax
\typeout{** WARNING: IEEEtran.bst: No hyphenation pattern has been}%
\typeout{** loaded for the language `#1'. Using the pattern for}%
\typeout{** the default language instead.}%
\else
\language=\csname l@#1\endcsname
\fi
#2}}
\providecommand{\BIBdecl}{\relax}
\BIBdecl

\bibitem{contreras2025probabilistic}
C.~A. Contreras, M.~Chiou, A.~Rastegarpanah, M.~Szulik, and R.~Stolkin, ``Probabilistic human intent prediction for mobile manipulation: An evaluation with human-inspired constraints,'' arXiv preprint arXiv:2507.10131, 2025.

\bibitem{ahn2022saycan}
M.~Ahn, A.~Brohan, N.~Brown, Y.~Chebotar, O.~Cortes, B.~David, C.~Finn, C.~Fu, K.~Gopalakrishnan, K.~Hausman \emph{et~al.}, ``Do as i can, not as i say: Grounding language in robotic affordances,'' \emph{arXiv preprint arXiv:2204.01691}, 2022.

\bibitem{pmlr-v229-zitkovich23a}
\BIBentryALTinterwordspacing
B.~Zitkovich, T.~Yu, S.~Xu, P.~Xu, T.~Xiao, F.~Xia, J.~Wu, P.~Wohlhart, S.~Welker, A.~Wahid, Q.~Vuong, V.~Vanhoucke, H.~Tran, R.~Soricut, A.~Singh, J.~Singh, P.~Sermanet, P.~R. Sanketi, G.~Salazar, M.~S. Ryoo, K.~Reymann, K.~Rao, K.~Pertsch, I.~Mordatch, H.~Michalewski, Y.~Lu, S.~Levine, L.~Lee, T.-W.~E. Lee, I.~Leal, Y.~Kuang, D.~Kalashnikov, R.~Julian, N.~J. Joshi, A.~Irpan, B.~Ichter, J.~Hsu, A.~Herzog, K.~Hausman, K.~Gopalakrishnan, C.~Fu, P.~Florence, C.~Finn, K.~A. Dubey, D.~Driess, T.~Ding, K.~M. Choromanski, X.~Chen, Y.~Chebotar, J.~Carbajal, N.~Brown, A.~Brohan, M.~G. Arenas, and K.~Han, ``Rt-2: Vision-language-action models transfer web knowledge to robotic control,'' in \emph{Proceedings of The 7th Conference on Robot Learning}, ser. Proceedings of Machine Learning Research, J.~Tan, M.~Toussaint, and K.~Darvish, Eds., vol. 229.\hskip 1em plus 0.5em minus 0.4em\relax PMLR, 06--09 Nov 2023, pp. 2165--2183. [Online]. Available: \url{https://proceedings.mlr.press/v229/zitkovich23a.html}
\BIBentrySTDinterwordspacing

\bibitem{gao2024physobjects}
J.~Gao, B.~Sarkar, F.~Xia, T.~Xiao, J.~Wu, B.~Ichter, A.~Majumdar, and D.~Sadigh, ``Physically grounded vision-language models for robotic manipulation,'' in \emph{2024 IEEE International Conference on Robotics and Automation (ICRA)}.\hskip 1em plus 0.5em minus 0.4em\relax IEEE, 2024, pp. 12\,462--12\,469.

\bibitem{huang2024lit}
Z.~Huang, J.~Pohovey, A.~Yammanuru, and K.~Driggs-Campbell, ``Lit: Large language model driven intention tracking for proactive human-robot collaboration--a robot sous-chef application,'' \emph{arXiv preprint arXiv:2406.13787}, 2024.

\bibitem{zhou2024lkirf}
J.~Zhou, X.~Su, W.~Fu, Y.~Lv, and B.~Liu, ``Enhancing intention prediction and interpretability in service robots with llm and kg,'' \emph{Scientific Reports}, vol.~14, no.~1, p. 26999, 2024.

\bibitem{liu2024vlm}
S.~Liu, J.~Zhang, R.~X. Gao, X.~V. Wang, and L.~Wang, ``Vision-language model-driven scene understanding and robotic object manipulation,'' in \emph{2024 IEEE 20th International Conference on Automation Science and Engineering (CASE)}.\hskip 1em plus 0.5em minus 0.4em\relax IEEE, 2024, pp. 21--26.

\bibitem{ultralytics2023yolov8}
Ultralytics, ``{YOLOv8} - you only look once,'' \url{https://github.com/ultralytics/ultralytics}, 2023, accessed: 2025-07-23.

\bibitem{kirillov2023segment}
A.~Kirillov, E.~Mintun, N.~Ravi, H.~Mao, L.~Rolland, M.~Schmidt, S.~Jenni, T.~Miller, S.~Graves, B.~Groth \emph{et~al.}, ``Segment anything,'' \emph{arXiv preprint arXiv:2304.02643}, 2023.

\end{thebibliography}

\end{document}